\documentclass[conference]{IEEEtran}
\IEEEoverridecommandlockouts
\usepackage{cite}
\usepackage{amsmath,amssymb,amsfonts}
\usepackage{algorithm} 
\usepackage{subcaption}
\usepackage{algpseudocode}
\usepackage{graphicx}
\usepackage{svg}
\usepackage{textcomp}
\usepackage{xcolor}
\usepackage{geometry}%
\usepackage{hyperref}
\usepackage{float}
\usepackage{eqexpl}
\usepackage{inputenc}
\DeclareUnicodeCharacter{202F}{!!!!FIXME!!!!}
\eqexplSetDelim{=}
\algnewcommand{\LineComment}[1]{\State \(\triangleright\) #1}
\algnewcommand{\parState}[1]{\State%
    \parbox[t]{\dimexpr\linewidth-\algmargin}{\strut\hangindent=\algorithmicindent \hangafter=1 #1\strut}}
\def\BibTeX{{\rm B\kern-.05em{\sc i\kern-.025em b}\kern-.08em
    T\kern-.1667em\lower.7ex\hbox{E}\kern-.125emX}}

\begin{document}

\title{Occupancy Grid Based Reactive Planner}

\author{
\IEEEauthorblockN{1\textsuperscript{st} Benjamin Hall}%
\IEEEauthorblockA{\textit{Robotics \& Mechatronics}\\
Marietta, United States\\
bhall90@students.kennesaw.edu}
\and
\IEEEauthorblockN{2\textsuperscript{nd} Andrew Goeden}
\IEEEauthorblockA{\textit{Computer Science}\\
Marietta, United States\\
agoeden@students.kennesaw.edu}
\and%
\IEEEauthorblockN{3\textsuperscript{rd} Sahan Reddy}
\IEEEauthorblockA{\textit{Computer Science}\\
Marietta, United States\\
sreddy13@students.kennesaw.edu}
\and%
\IEEEauthorblockN{4\textsuperscript{th} Timothy Gallion}
\IEEEauthorblockA{\textit{Electrical Engineering}\\
Marietta, United States\\
timgallion315@gmail.com}
\and%
\IEEEauthorblockN{5\textsuperscript{th} Charles Koduru}
\IEEEauthorblockA{\textit{Robotics \& Mechatronics}\\
Marietta, USA\\
ckoduru@students.kennesaw.edu}
\and%
\IEEEauthorblockN{6\textsuperscript{th} M. Hassan Tanveer}
\IEEEauthorblockA{\textit{Robotics \& Mechatronics}\\
Marietta, USA\\
mtanveer@kennesaw.edu}
}
\maketitle

\begin{abstract}
This paper proposes a perception and path planning pipeline for autonomous racing in an unknown bounded course. The pipeline was initially created for the 2021 evGrandPrix autonomous division and was further improved for the 2022 event, both of which resulting in first place finishes. Using a simple LiDAR-based perception pipeline feeding into an occupancy grid based expansion algorithm, we determine a goal point to drive. This pipeline successfully achieved reliable and consistent laps in addition with occupancy grid algorithm to know the ways around a cone-defined track with an averaging speeds of 6.85 m/s over a distance 434.2 meters for a total lap time of 63.4 seconds.
\end{abstract}


\section{Introduction}
Since vehicles have existed, they have been raced against one another. Racing has been the breeding ground for many innovations in the automotive industry that we take for granted today, such as more efficient engines, lighter and stronger materials, and safety elements\cite{Young2012FormulaOR}. Autonomous racing seeks to replicate this success of traditional racing in contributing to driven vehicles but instead to the field of autonomy\cite{Alexander2021AutonomousRace}. Autonomous racing, generally, can be defined as either a head-to-head or fastest solo lap around a track by a vehicle that is controlling itself\cite{Betz2022Survey}.

\begin{figure}
    \centering
    \includegraphics[width=\columnwidth]{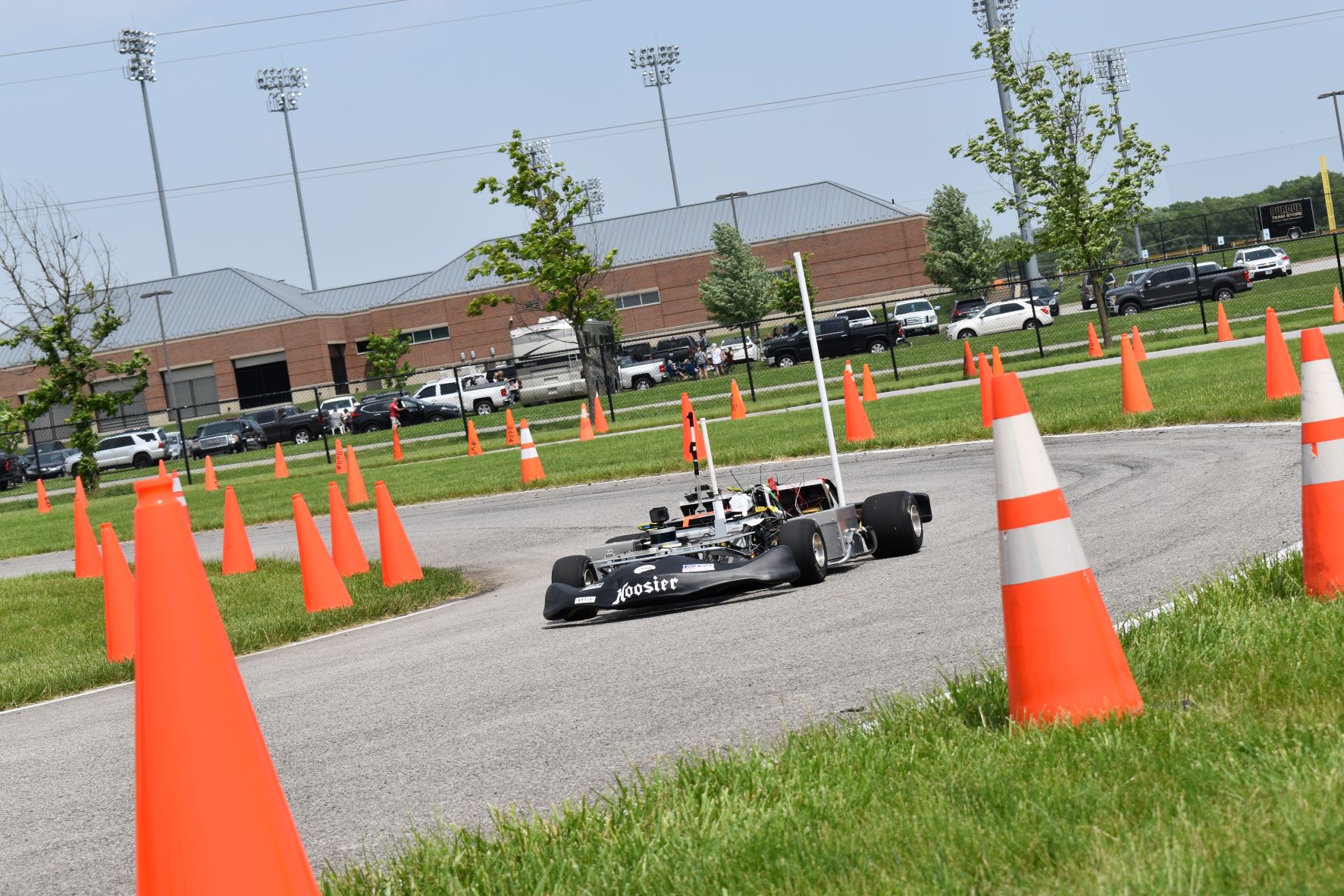}
    \caption{EVT Autonomous Go-Kart Driving On Track}
    \label{fig:voltron_driving}
\end{figure}

Our competition is based on how fast an autonomous go-kart can go around a cone-defined track. The kart cannot start with a map of the track, but mapping can be done throughout the race\cite{evgrand_prix_2021}. This paper focuses on the reactive planner that is used during the first lap of the race before we have mapped out the track. The approaches to solve the challenges in this competition, such as the first lap track exploration problem, were inspired by solutions from other competitions such as F1TENTH and Formula Student Driverless.

The Formula Student Driverless competition contains the similar problem of navigating an unknown track. Within Formula Student Driverless, the track is defined by two sides of different colored cones, each color representing a different side of the track. Within F1tenth, the track is defined by continuous walls. F1tenth has solved the problem of navigation around the track with approaches such as follow-the-gap \cite{disparity_extender}. Within Formula Student Driverless, solutions most often include cone detection and classification and since the inside and outside of the track are known at all points along the track the solution for determining a path to navigate within becomes trivial \cite{Andresen2020AccurateMA}. Within evGrandPrix, the track is defined by semi-continuous walls of the same color cone\cite{evgrand_prix_2021}.

Compared to many Formula Student Driverless solutions to the problem, our approach does not require object classification or a pre-defined inside and outside track \cite{Andresen2020AccurateMA}. We also do not generate many different possibles paths each time like a Monte-Carlo approach, but instead create a single path each time \cite{Jonsson_Stenback_2020}. Similar to follow-the-gap our planning stage requires a continuous track edge definition, however our perception stage produces a simulated false wall. Our planning and perception stages are loosely coupled, with the only constraint being that the perception stage must output a continuous wall on at least one side of the track at all times.

\section{Autonomous Driving Pipeline}

\subsection{Overview}
Our proposed autonomous driving pipeline uses a local perception, planning and control strategy to drive around a bounded track. In our context, local refers to the coordinate system with its origin located on the kart itself. The perception pipeline is expected to create an occupancy grid with continuous boundaries on the left and right of the kart. Then, the path planner selects a goal point on this occupancy grid, which is driven to with our pure pursuit controller. The pipeline was implemented leveraging ROS 2 Galactic \cite{Ros2Galactic} along with the Point Cloud Library\cite{Rusu_ICRA2011_PCL}, OpenCV \cite{opencv_library} and Eigen \cite{eigenweb}.

\begin{figure}
    \centering
    \includegraphics[width=\columnwidth]{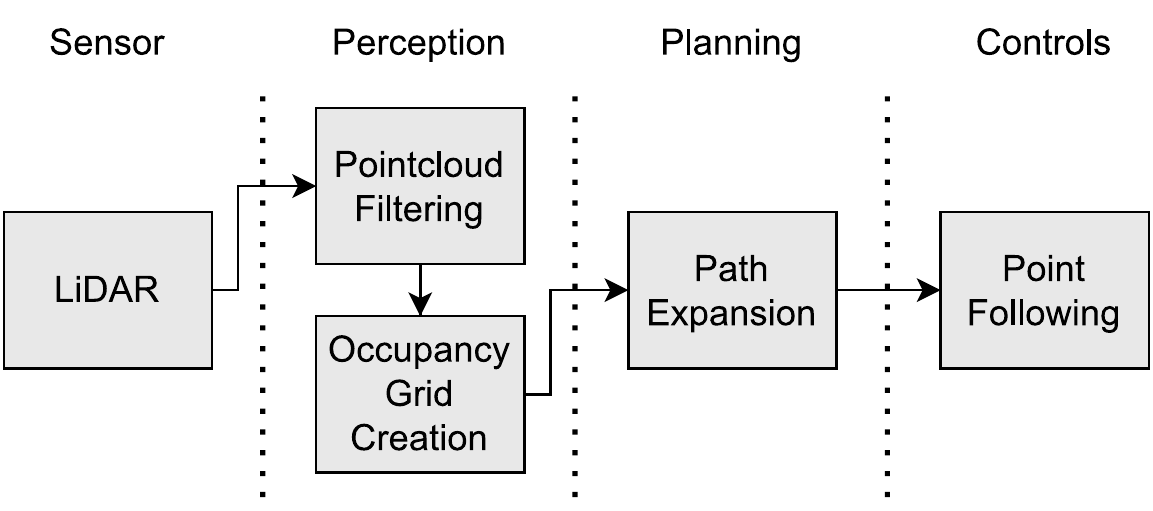}
    \caption{Pipeline Overview}
    \label{fig:pipeline_overview}
\end{figure}

\subsection{Perception}

\begin{figure}
    \centering
    \includegraphics[width=\columnwidth]{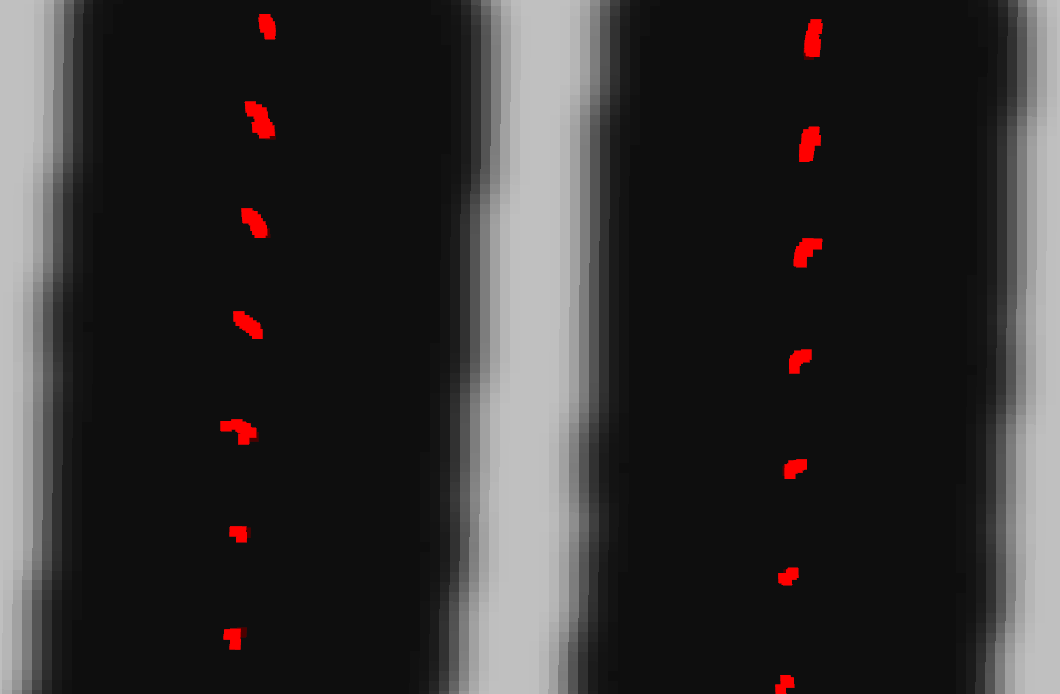}
    \caption{Blurred Occupancy Grid}
    \label{fig:blurred_costmap}
\end{figure}

 In the evGrandPrix competition, the track is defined by traffic cones spaced approximately 2 meters apart. To perceive this track boundary, we take an entirely LiDAR-based approach with a Velodyne VLP-16 \cite{velodyne_lidar}. First, the ground is segmented out of the pointcloud using RANSAC \cite{RANSAC}---we assume that any points left over are obstacles that define our boundary. These non-ground points are flattened into an occupancy grid. Finally, this occupancy grid is inflated with a Gaussian blur in order to (1) create a continuous wall from the discrete cones and (2) create a "cost valley" in the center of the track so that the planner follows a rough centerline. The result of this can be seen in Fig. \ref{fig:blurred_costmap}.

\subsection{Path Planning}

\begin{figure}[ht]
    \centering
    \begin{subfigure}{.49\columnwidth}
        \includegraphics[width=\columnwidth]{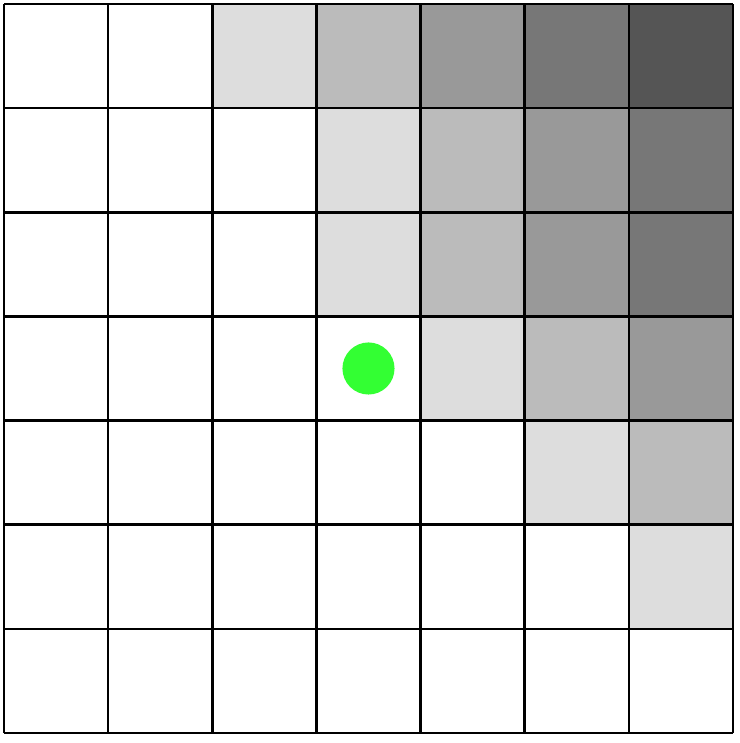}
        \caption{Initial}
        \label{fig:pipeline_expansion_initial}
    \end{subfigure}
    \hfill
    \begin{subfigure}{.49\columnwidth}
        \includegraphics[width=\columnwidth]{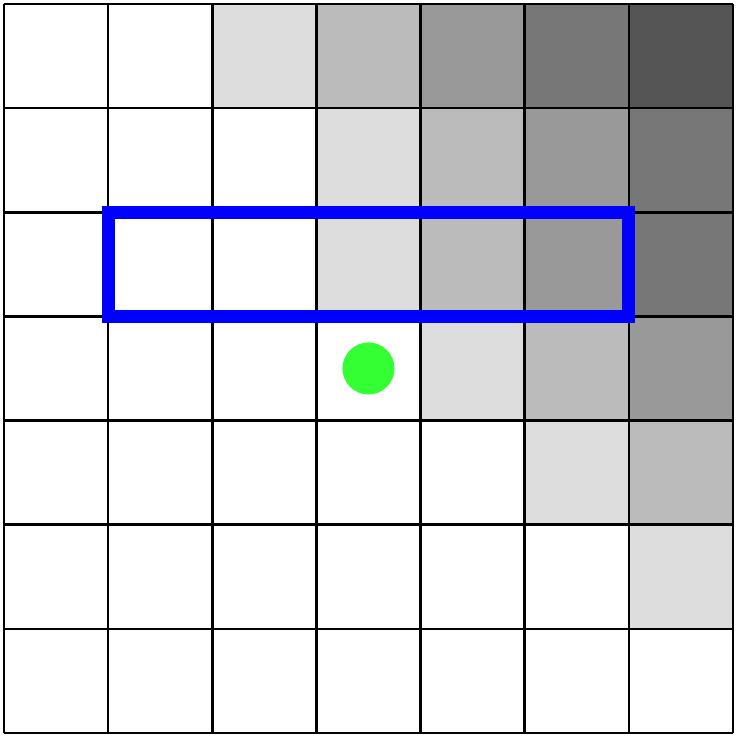}
        \caption{Expansion space}
        \label{fig:pipeline_expansion_subspace}
    \end{subfigure}
    \begin{subfigure}{.49\columnwidth}
        \includegraphics[width=\columnwidth]{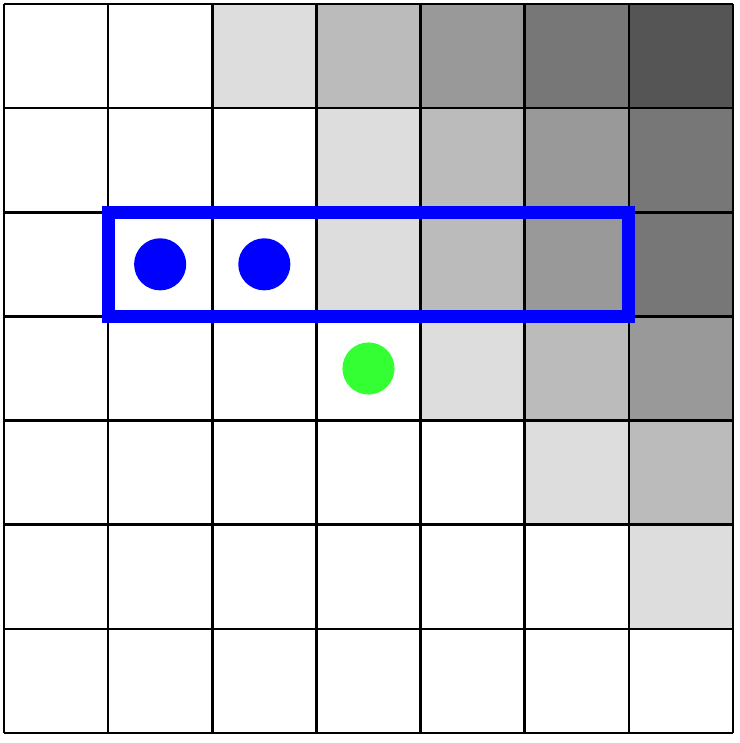}
        \caption{Identify cells with the least cost}
        \label{fig:pipeline_expansion_least_cost}
    \end{subfigure}
    \begin{subfigure}{.49\columnwidth}
        \includegraphics[width=\columnwidth]{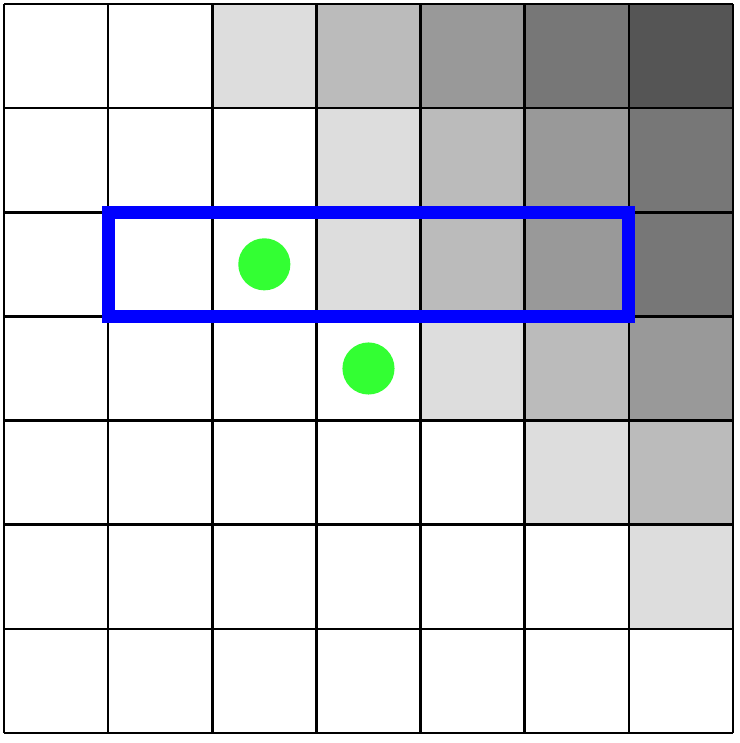}
        \caption{Select the one with the least deviation}
        \label{fig:pipeline_expansion_least_deviation}
    \end{subfigure}
    \begin{subfigure}{.49\columnwidth}
        \includegraphics[width=\columnwidth]{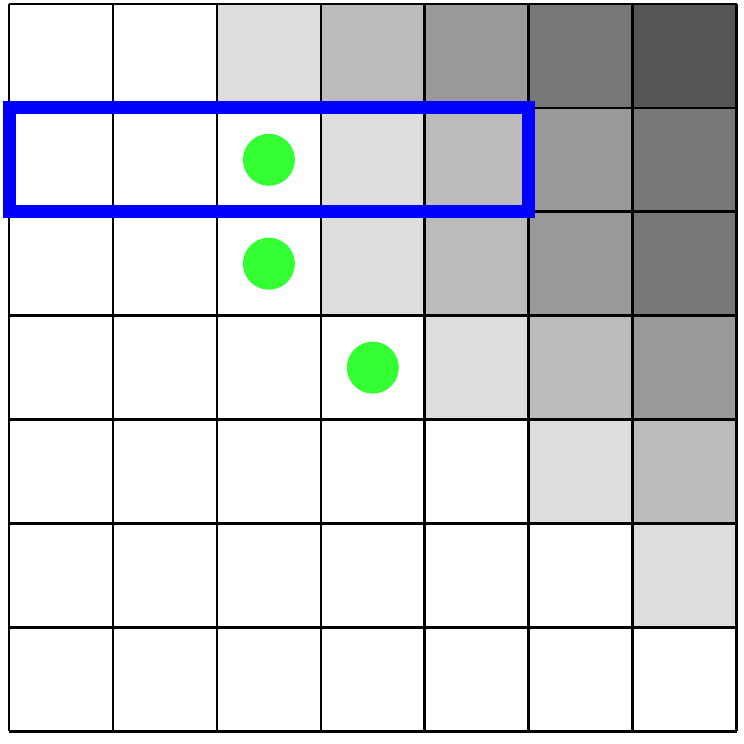}
        \caption{Repeat until maximum expansions is reached}
        \label{fig:pipeline_expansion_repeat}
    \end{subfigure}
    \caption{The pipeline scans each row and identifies the spaces with least cost and then selects the space of least cost with the least deviation before repeating on the next row.}
    \label{fig:pipeline_expansion_steps}
\end{figure}

Once we have created an occupancy grid that has continuous walls, we can start to plan our path. Our planning algorithm (Algorithm \ref{alg:path_planning}) works by completing a set number of expansions starting from an initial node. This node is at the location of the LiDAR, which is the center of the grid (Fig. \ref{fig:pipeline_expansion_initial}).

The structure of each expansion follows suit: we first define the subspace (row) ahead of the LiDAR in which we plan to scan for least cost (Fig. \ref{fig:pipeline_expansion_subspace}). This subspace is defined by $check\_width$ which is centered from the previous node (Fig. \ref{fig:pipeline_expansion_least_cost}). After finding the points with the least costs, if there are multiple with the same cost then we select the point with the least deviation from the previous/current node (Fig. \ref{fig:pipeline_expansion_least_deviation}) to be the next node in the path, otherwise we select the singular point of least cost as the next node (Fig. \ref{fig:pipeline_expansion_repeat}).

The path creation is done via a point-by-point expansion from the center of the grid which is the location of the go-kart. The zone we are allowed to expand to is above and an amount to the left and right of this initial point. Then the cell inside of this zone that we expand to is chosen in two steps: first, whichever one has least cost and second if there is a tie between multiple cells, the cell closest to the center of the expansion zone, and therefore the cell we're expanding from, is chosen.

This expansion process repeats a set number of times, $n$. The last point in this series of expansions is chosen as our goal point to drive to. This process can be seen in figure \ref{fig:pipeline_expansion_steps}. The algorithm that describes this can be seen in Algorithm \ref{alg:path_planning}.

We do this using pure pursuit \cite{Coulter1992ImplementationOT}, which controls our steering angle while driving forward at a constant velocity. 

\begin{algorithm}[ht]
\caption{Path Planning Algorithm}\label{alg:path_planning}
\begin{algorithmic}
\Require Occupancy Grid
\Ensure Goal Point
\State $x \gets grid\_width / 2$, current node x
\State $y \gets grid\_height / 2$, current node y

\State $half\_width$, half the configured width of the expansion space
\For{$n \gets 0$ to $max\_expansions$}
    \State $x'$
    \State $y' \gets y + 1$
    \State $least\_cost \gets \infty$
    \State $least\_x\_deviation \gets \infty$
    \For{$c \gets -half\_width$ to $half\_width$}
        \State {$expand\_x$ $\gets$ $c + x$}
        \If{$grid[y'][expand\_x] \leq least\_cost$ \textbf{and} $|x - expand\_x| < least\_x\_deviation$}
            \State $least\_cost \gets grid[y'][expand\_x]$
            \State $least\_x\_deviation$%
            $\gets |x - expand\_x|$
            \State $x' \gets expand\_x$
        \EndIf
    \EndFor

    \State $x \gets x'$
    \State $y \gets y'$
\EndFor
\end{algorithmic}
\end{algorithm}

\section{Results and Discussion}

Using this pipeline, the kart was able to fully navigate the track with surprising consistency: over the five laps of the race, the completion time variance was within a quarter second. For these laps, our average speed was 6.85 m/s; there is much room for improvement here, since our pure pursuit goal point follower runs at a constant speed.  

Because the planner has to move forward in the occupancy grid with every expansion and it has a limited space that it can expand into, in a case where the only options in the expansion space are further into a high cost boundary, it will choose a goal point past that boundary. This means that it will fail when presented with a wall perfectly perpendicular to the kart or one in an upside down "U" shape in front of the kart. For our use case, we can assume that this happens extremely rarely, since a track has smooth curves. 

One possible improvement to this pipeline is addition that could tune the occupancy grid blur dynamically to always produce an optimal "cost valley" so that the planner plans along a rough center line. This would greatly reduce manual tuning time and would improve performance on width-varying tracks. In addition, this planner could potentially be used with a 3D occupancy grid; a use case for this is planning a goal point for a drone to fly around terrain

\begin{figure}
    \centering
    \includegraphics[width=\columnwidth]{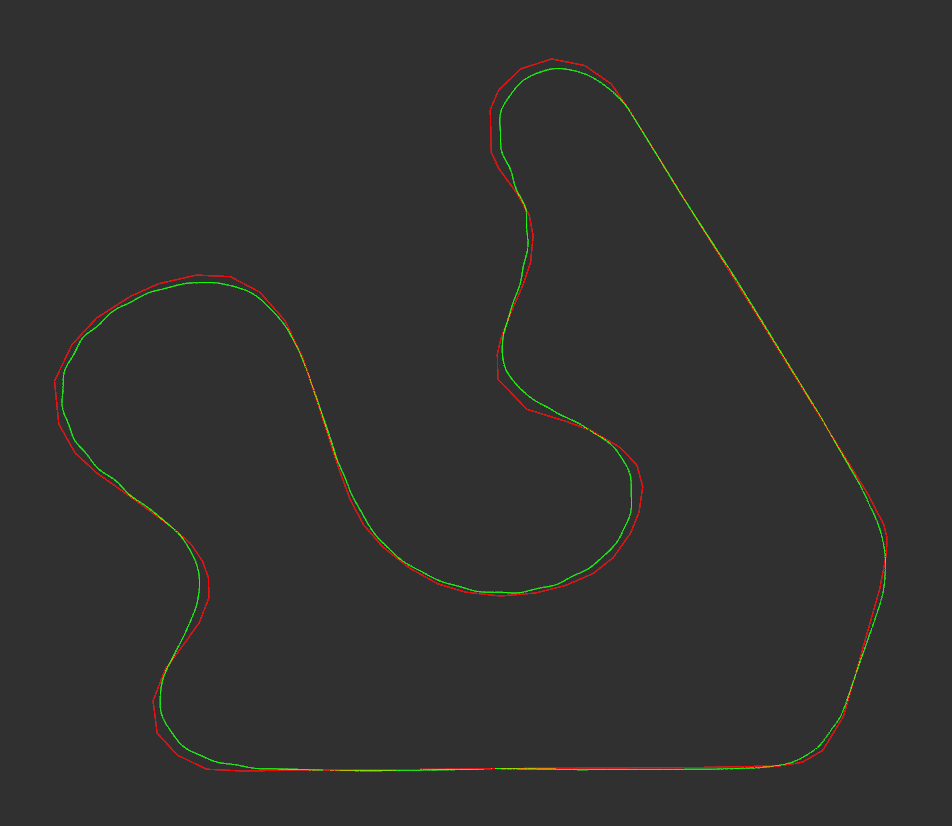}
    \caption{The path of the kart (green) over the true track centerline (red).}
    \label{fig:track_run}
\end{figure}
 
\section{Conclusion}
This paper presented a solution to the problem of traversing an unknown cone defined race track within Autonomous racing. Our solution is beneficial for the whole of autonomous racing because it provides a more simplistic and generalized approach to navigating an unknown track. The algorithm presented shows that utilizing an occupancy grid based method to navigate a race track is feasible and robust. Additionally, the pipeline implemented here can be extended further into more optimized solutions that can improve on what we have shown. We are excited to see further developments and optimizations from utilizing this pipeline within other existing and new autonomous racing pipelines. The code for this algorithm can be found \href{https://gitlab.com/KSU_EVT/autonomous-software/pointcloud-to-costmap}{here}.

\section{Acknowledgements}
We would like to thank Kennesaw State University Electric Vehicle Team for providing the opportunities, resources, and support to come up with this pipeline. We would also like to thank the evGrandPrix organization and Purdue University for their gracious dedication to this competition, in addition to the other participating universities for their constant effort in growing and shaping the community and challenges associated with this new competition.

\bibliographystyle{IEEEtran}
\bibliography{references}

\end{document}